\documentclass[letterpaper, 10 pt, journal, twoside]{IEEEtran}
\IEEEoverridecommandlockouts

\usepackage{graphics}
\usepackage{pgfplots}
\usepackage{epsfig}
\usepackage{amsmath}
\usepackage{amssymb}
\usepackage{multirow}
\usepackage{tabularx}
\usepackage{booktabs}
\usepackage{subcaption}
\usepackage{verbatim}
\usepackage{listings}
\usepackage{xcolor} 
\usepackage{amrl}
\let\labelindent\relax
\usepackage[inline]{enumitem}
\newif\ifdraft
\drafttrue

\usepackage{pervasives}
\usepackage{hyperref}
\usepackage[noadjust]{cite}

\newcolumntype{C}{>{\centering\arraybackslash}X} 

\begin{document}

\title{\LARGE \bf
Deploying and Evaluating LLMs to Program Service Mobile Robots
}

\author{Zichao Hu$^{1}$, Francesca Lucchetti$^{2}$, Claire Schlesinger$^{2}$, Yash Saxena$^{1}$, 
Anders Freeman$^{3}$, \\ Sadanand Modak$^{1}$, Arjun Guha$^{2}$,  Joydeep Biswas$^{1}$
\thanks{This work is partially supported by the National Science Foundation (CCF-2006404 and CCF-2102291) and JP Morgan. Any opinions, findings, and conclusions expressed in this material are those of the authors and do not necessarily reflect the views of the sponsors.}
\thanks{$^{1}$Department of Computer Science, University of Texas at Austin, \{zichao, yash.saxena, sadanandm, joydeepb\}@utexas.edu}%
\thanks{$^{2}$Khoury College of Computer Sciences, Northeastern University, \{lucchetti.f, schlesinger.e, a.guha\}@northeastern.edu}%
\thanks{$^{3}$Department of Computer Science, Wellesley College, af103@wellesley.edu}%

}

\maketitle


\begin{abstract}
Recent advancements in large language models (LLMs) have spurred interest in using them for generating robot programs from natural language, with promising initial results.  We investigate the use of LLMs to generate programs for service mobile robots leveraging mobility, perception, and human interaction skills, and where \emph{accurate sequencing and ordering} of actions is crucial for success. We contribute \codebotler{}, an open-source robot-agnostic tool to program service mobile robots from natural language, and \roboeval{}, a benchmark for evaluating LLMs’ capabilities of generating programs to complete service robot tasks. \codebotler{} performs program generation via few-shot prompting of LLMs with an embedded domain-specific language (eDSL) in Python, and leverages skill abstractions to deploy generated programs on any general-purpose mobile robot.  \roboeval{} evaluates the correctness of generated programs by checking execution traces starting with multiple initial states, and checking whether the traces satisfy temporal logic properties that encode correctness for each task.  \roboeval{} also includes multiple prompts per task to test for the robustness of program generation. We evaluate several popular state-of-the-art LLMs with the \roboeval{} benchmark, and perform a thorough analysis of the modes of failures, resulting in a taxonomy that highlights common pitfalls of LLMs at generating robot programs. 
We release our code and benchmark at
\href{https://amrl.cs.utexas.edu/codebotler/} {https://amrl.cs.utexas.edu/codebotler/}.

\end{abstract}


\section{Introduction}

\begin{figure*}[h!]
    \centering
    \includegraphics[width=\textwidth]{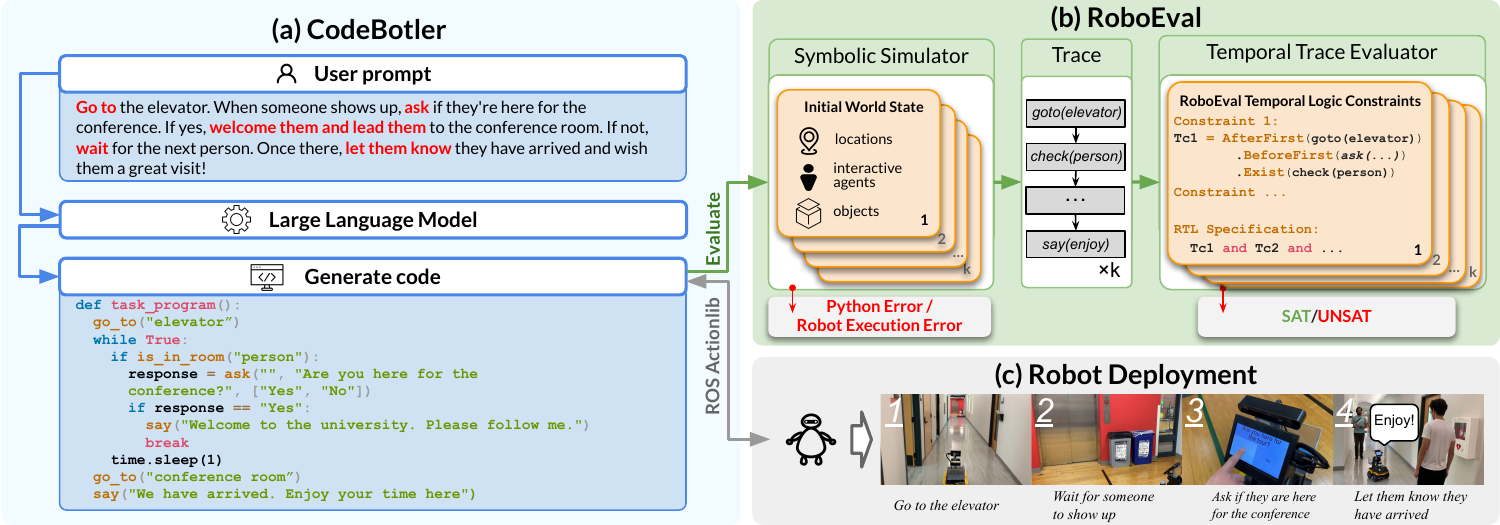}
    \caption{The system diagram of \codebotler{} and \roboeval{}. \codebotler{} receives a user prompt and queries a large language model (LLM) to generate a robot program (\textbf{a}). It can execute the program to send instructions to a robot via ROS Actionlib (\textbf{c}). Separately, \roboeval{} evaluates programs generated from its benchmark tasks using a symbolic simulator and a temporal trace evaluator to determine whether each program satisfies the task constraints or not (\textbf{b}).}
    \label{fig:system-diagram}
    \figspace
\end{figure*}

We are interested in deploying service mobile robots to perform arbitrary user tasks from natural language descriptions. Recent advancements in large language models (LLMs) have
shown promise in related applications involving visuomotor tasks~\cite{codeaspolicies2022,singh2023progprompt}, planning~\cite{ahn2022isaycan,huang2022language, driess2023palme, liu2023llmp}, and in this paper, we investigate the use of LLMs to generate programs for \emph{service
mobile robots} leveraging mobility, perception, and human interaction skills,
 where accurate \emph{sequencing and ordering of actions} is crucial for success. 
 We contribute \codebotler{} and \roboeval{}: \codebotler{} is an open-source robot-agnostic tool to generate general-purpose service robot programs from natural language, and \roboeval{} is a benchmark for evaluating LLMs’ capabilities of generating programs to complete service robot tasks.

\codebotler{} leverages an embedded domain-specific language (eDSL) in Python to abstract key robot skills, and includes robot-agnostic bindings for such tasks using ROS Actions~\cite{rosactionlib}. Given few-shot examples, \codebotler{} uses LLMs to convert natural language task descriptions into programs in the eDSL, which are then executed on real robots using a lightweight interpreter to interface with robot-specific skills. 

While the capabilities of LLMs at producing robot programs are impressive, they are still susceptible to a variety of failures. To understand empirically the failure modes of LLMs in producing robot programs, we introduce the \roboeval{} benchmark. Given a program generated by \codebotler{}, \roboeval{} first executes it in a symbolic simulator to generate multiple program traces from different initial world states, and then checks these traces against a set of temporal checks that define correct behavior for the task for each initial world state.

Existing code completion benchmarks tackle simple data
processing functions~\cite{humaneval} or low-level robot skills~\cite{codeaspolicies2022}, which are amenable to simple input-output unit tests. However, code generation for general-purpose service robot programs cannot be
evaluated just on input/output sequences. For example, given a task
\emph{``Check how many conference rooms have no markers''}, it is insufficient
to just test whether the LLM-generated program executes to state the correct
answer --- to ensure correctness, a correct program must first visit all
conference rooms and check for markers there before arriving at the result.
We also 
observe that there are significant variations in the correctness of
generated programs with small variations in the phrasing of the natural language
task descriptions~\cite{studenteval}. We thus contribute three key ideas to test for both
correctness and robustness of LLM-generated robot programs: 1) we evaluate the
\emph{execution traces} of programs; 2) we check
whether the execution traces satisfy \emph{temporal logic} properties that
encode correctness for each task; and 3) we \emph{vary the prompts} and
to test for robustness. \figref{system-diagram} shows the system diagram of \codebotler{} and \roboeval{}.

We further categorize the types of failures of different LLMs in the \roboeval{} benchmark and find several common categories of failures, including Python run-time errors, errors in executing infeasible robot actions, and errors in satisfying task requirements. We analyze the types of errors in each category and find several common modes of failures across LLMs. We believe this analysis will be invaluable in furthering research on LLM-guided robot program generation. Driven by our initial findings, we include a simple rejection sampling procedure that shows immediate improvements in reducing robot execution errors of LLM-generated robot programs.

In summary, this paper contributes:
\begin{enumerate}
    \item \codebotler{}, an open-source tool to generate robot programs from natural language using LLMs, and to enable robot-agnostic deployment of such programs;
    \item \roboeval{}, a benchmark to evaluate LLM-generated robot programs for service mobile robots;
    \item a comprehensive analysis and taxonomy of failures of LLM-generated robot programs; and
    \item a rejection sampling mechanism to reduce robot execution failures of LLM-generated robot programs.
\end{enumerate}

\section{Related Work}

 \emph{\textbf{LLMs for Robotics Applications.}}
 Using LLMs to perform robotics tasks~\cite{huang2023vlmap, ding2023task, wang2023demo2code, huang2023voxposer, huang2022language, liu2023llmp, ahn2022isaycan, driess2023palme, brohan2023rt2, tang2023saytap, yu2023language, codeaspolicies2022, singh2023progprompt, shah2022lmnav, huang2022inner, wu2023tidybot, chen2022openvocabulary, song2023llmplanner} has attracted a lot of attention because of their impressive out-of-box and commonsense reasoning capabilities~\cite{brown2020language}. One approach uses LLMs as task planners to break down a free-form natural language task description into multiple sub-goals. Language Models as Zero-Shot Planners~\cite{huang2022language} expresses these sub-goals in the form of natural language and builds an interpreter to convert these subgoals into robot actions. LLM+p~\cite{liu2023llmp} outputs these sub-goals in the form of the well-defined planning domain definition language (PDDL). Another approach leverages the code-writing capabilities of LLMs to generate programs for the robot to execute. Code-as-Policies~\cite{codeaspolicies2022} defines a robot-centric formulation of language model-generated programs (LMPs). It proposes a hierarchical method to query an LLM and generate executable Python programs that invoke parameterized robot primitives. Voxposer~\cite{huang2023voxposer} builds on the Code-as-Policies' LMP formulation and defines a set of primitives that enables the LLMs to generate Python programs to create voxel cost maps. Then, it plans on the voxel cost maps and carries out the specified manipulation tasks. \codebotler{} builds on the LMP formulation, in which we define a set of 8 robot primitives specific to service mobile robots. 

\emph{\textbf{Evaluating LLMs.}}
The evaluation of LLMs is an ongoing effort due to the scope and variability of model generations~\cite{chang2023survey, liang2023holistic, liu2023agentbench}. Much progress has been made in evaluating code-writing LLMs~\cite{studenteval, humaneval, lai2022ds1000, austin2021program}. While these programming tasks assess language understanding, reasoning, algorithms, and basic mathematics, they do not address the skills of embodied agents. In the domain of embodied agents, ProgPrompt~\cite{singh2023progprompt} creates a dataset of 70 household high-fidelity 3D simulation tasks to evaluate the code-writing capabilities of LLMs. High-fidelity 3D simulations are useful for capturing complex agent-environment dynamics but not essential for verifying program logic, and creating them can be time-consuming. To address this limitation, we create \roboeval{}, a lightweight benchmark that uses a \emph{symbolic} simulator to evaluate the temporal correctness of a robot program.

\section{\codebotler{}: Robot-Agnostic Deployments}
\codebotler{} is a lightweight open-source tool that 1) defines robot-agnostic skills; 2) provides a user-friendly web interface to accept instructions and generate language model programs (LMPs) using LLMs; and 3) executes LMPs independently of robot platforms by sending commands to robots via ROS Actionlib. To illustrate the capabilities of \codebotler{}, we present its design in the following subsections.

\begin{figure*}[h]
    \includegraphics[width=\textwidth]{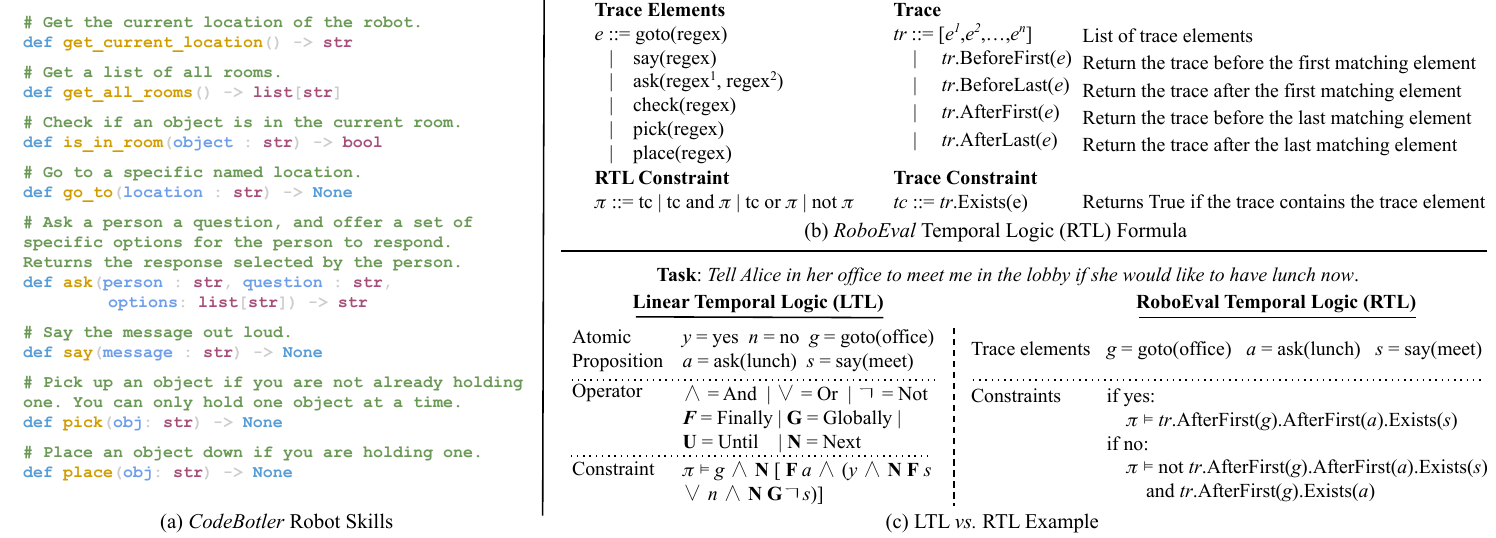}
    \caption{\codebotler{} robot skills (\textbf{a}), \roboeval{} temporal logic (RTL) formula (\textbf{b}), and the LTL specifications \vs{} the RTL specifications of an example task (\textbf{c}). In section (\textbf{c}), the terms \emph{office}, \emph{meet}, and \emph{lunch} are used to represent the regex patterns. The RTL specifications are simpler to express and have improved readability.}
    \label{fig:codebotler-roboeval}
    \figspace
\end{figure*}

\subsection{\codebotler{} Language Design}

\codebotler{} leverages an embedded domain-specific language (eDSL) in Python to abstract 8 commonly available service robot skills. \subfigref{codebotler-roboeval}{a} shows the definition of each skill. The design of the \codebotler{} eDSL encompasses skills such as:
\begin{enumerate*}
    \item \texttt{get\_current\_location} and \texttt{get\_all\_rooms} that inspect the robot's state and the world configurations, so they do not need to be hard-coded or manually specified in the prompt, as in previous work~\cite{codeaspolicies2022, huang2023voxposer, singh2023progprompt}, 
    \item \texttt{is\_in\_room} that utilizes the zero-shot visual-language models (VLMs) for perceptual reasoning,
    \item \texttt{ask} that provides a structured interface for human interaction through multiple-choice questions, facilitating both the LMP in processing human responses and interaction with a robot via touch-screen or audio input.
    \item \texttt{pick}, \texttt{place}, and \texttt{go\_to} that represent core robot manipulation and navigation abilities.
\end{enumerate*}
 These abstractions allow \codebotler{} to be used for robot-agnostic deployments. An LMP can often be reused across different maps and robots through the user interface. 

\subsection{User Interface (UI) and Robot Program Generation}
The \codebotler{} UI includes a task input pane, a program preview pane, and a status monitor to track the generation of programs on the robot.
When given a user task, \codebotler{} combines it with a prompt prefix containing robot skills and a few example programs, and then queries an LLM for program generation. In addition, \codebotler{} supports many LLM interfaces, including the OpenAI API~\cite{OpenAIPlatform}, the Google PaLM API~\cite{PaLMAPI}, and HuggingFace models (\texttt{AutoModel} and \texttt{Text-Generation-Inference}).

\subsection{Robot Deployment With \codebotler{}}
\codebotler{} is designed to work with the ROS system and acts as a client by utilizing the ROS Actionlib~\cite{rosactionlib}. When \codebotler{} executes an LMP and encounters a statement that invokes a robot skill (\eg{} \texttt{go\_to("lobby")}), it publishes a goal (\eg{} \texttt{"lobby"}) to the appropriate remote topic (\eg{} \texttt{"/go\_to\_server"}) for a ROS action server on the robot to pick up. This approach makes \codebotler{} independent of any specific robot platform and permits \codebotler{} to operate both onboard and externally to the robot. Additionally, it provides robot deployers with the flexibility to customize the ROS action server to accommodate their specific needs for these primitives\footnote{The code for our implementation of the robot skills can be found at \href{https://github.com/ut-amrl/codebotler_amrl_impl}{https://github.com/ut-amrl/codebotler\_amrl\_impl}.}.

\begin{figure*}[t]
    \centering
    \includegraphics[width=\textwidth]{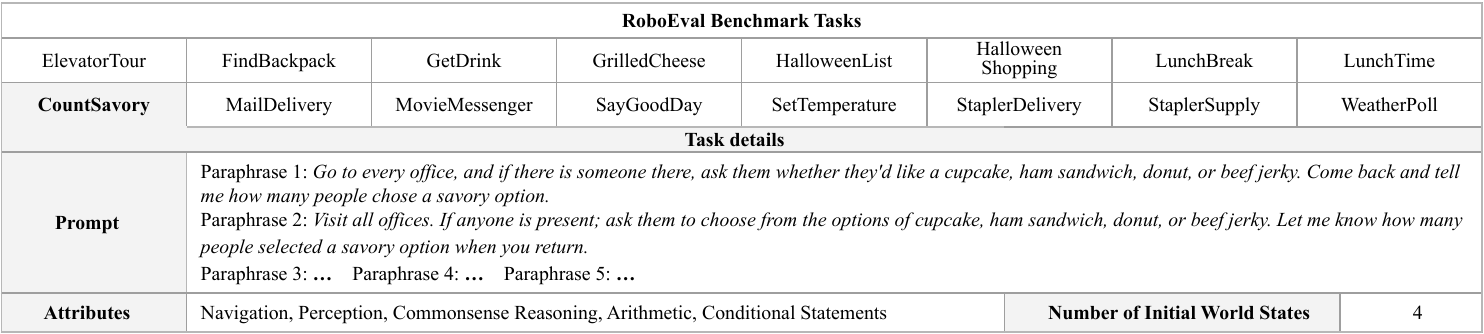}
    \caption{The \roboeval{} benchmark includes 16 tasks, each with 5 prompt paraphrases. The figure displays these tasks' names and a detailed example of the task \texttt{CountSavory}.}
    \label{fig:benchmark-tasks}
    \figspace
\end{figure*}


\section{The \roboeval{} Benchmark}
\newcommand{\Finally}{\ensuremath{\mathcal{F}}}
\newcommand{\Globally}{\ensuremath{\mathcal{G}}}
\newcommand{\Until}{\ensuremath{\mathcal{U}}}
\newcommand{\Next}{\ensuremath{\mathcal{N}}}

\roboeval{} consists of a simulator, an evaluator, and a benchmark suite of tasks. Given $P$, the space of natural language prompts describing service mobile tasks, and $\Pi$, the set of possible LMPs, \codebotler{} generates LMPs $\pi \in \Pi$ given a prompt $p \in P$. The symbolic simulator accepts a world state $w\in W$ and an LMP $\pi \in \Pi$, and produces a program trace $r\in R$. The evaluator accepts a trace and a temporal constraint $c\in C$, and returns whether the trace satisfies the constraint or not (\texttt{SAT}/\texttt{UNSAT}).
\begin{align*}
    \codebotler{}: P & \rightarrow \Pi \\
    \mathrm{Simulator}: \Pi \times W &\rightarrow R \\
    \mathrm{Evaluator}: R \times C &\rightarrow \{\texttt{SAT}, \texttt{UNSAT}\}
\end{align*}
The results derived from traces over multiple world states and multiple task prompts yield the success rate for an LLM on a particular task. The \roboeval{} benchmark thus consists of tasks $T_i (i\in [1,N])$, where each task consists of $M$ prompts, and $K$ world states. Each world state has a corresponding temporal check. Each \roboeval{} task thus consists of a tuple of prompts and multiple world-states to check against a constraint (one constraint per world state):
\begin{align*}
    T_i = \left\langle \{p_i^j| j\in[1,M]\}, \{\langle w_i^k, c_i^k\rangle| k\in[1,K] \} \right\rangle
\end{align*}

We present next
\begin{enumerate*}
    \item the \roboeval{} simulator, 
    \item the \roboeval{} evaluator, and 
    \item the tasks in the \roboeval{} benchmark.
\end{enumerate*}

\subsection{The \roboeval{} Simulator}
For each task $T_i$, the \roboeval{} benchmark includes multiple world states to check against. Each world state $w_i^k\in W$ consists of
\begin{enumerate*}
    \item a list of rooms in the world that \texttt{GetAllRooms()} returns, and which \texttt{GoTo()} is valid for;
    \item a list of objects in the world that \texttt{IsInRoom()} returns true for; 
    \item a list of objects that can be manipulated using \texttt{Pick()} and \texttt{Place()}; and
    \item a list of responsive humans, their locations, and regular expressions that define their responses to \texttt{Ask()}.
\end{enumerate*}
Thus, a single LMP may produce very different traces when simulated with different initial world states. The simulator consists of a Python interpreter and a symbolic simulation of each robot skill, and the result of running an LMP $\pi$ is recorded as a trace $r$ as a sequence of robot skills that were executed, along with the parameters (\eg{} the location parameter of a \texttt{GoTo} call). All Python errors or robot execution errors are logged during simulation.

\subsection{The \roboeval{} Evaluator}
Given a trace $r_i^k$ produced by simulating an LMP $\pi$ with an initial world state $w_i^k$, the \roboeval{} evaluator checks whether $r_i^k$ satisfies the temporal check $c_i^k$ that defines correct execution of the task for that world state. $c_i^k$ may consist of multiple conditions, expressed in conjunctive normal form over multiple temporal constraints. We review Linear Temporal Logic, which is well-suited to codifying such constraints in order to check for correctness.

\emph{\textbf{Linear Temporal Logic.}}
An LTL formula follows the grammar shown in \subfigref{codebotler-roboeval}{c} --- it composes atomic propositions $\pi\in \Pi$ with logical operators $\neg,\wedge,\vee$ and temporal operators $\Finally,\Globally,\Until,\Next$. Given LTL formulas $\phi_1,\phi_2$ defined over a temporal sequence, $\Finally \phi_1$ is true iff $\phi_1$ is true eventually at some point along the sequence, $\Globally \phi_1$ is true iff $\phi_1$ is true over the entire sequence, and $\phi_1 \Until \phi_2$ is true iff $\phi_1$ for a sub-sequence and $\phi_2$ is true for the remainder of the sequence after that. $\Next \phi$ is true for a sequence iff the next element in the sequence satisfies $\phi$.

\emph{\textbf{\roboeval{} Temporal Logic}}
While LTL suffices for writing robot task specifications, these LTL formulas can become complex as task complexity increases.
For example, consider an example task $\mathcal{T}_1$ where a user asks the robot,  \emph{``tell Alice in her office to meet me in the lobby if she agrees to lunch"}. To complete this task, the robot 1) \emph{first} needs to go to Alice's office; 2) \emph{then} ask Alice whether she would like to have lunch; and 3) \emph{finally} if she agrees, tell her to meet in the lobby.
\subfigref{codebotler-roboeval}{c} shows the complete LTL specification for this task. 
Declaring such specifications is quite tedious and error-prone.
To address this challenge, we observe that 1) specifying temporal logic is easier and less error-prone for specific scenarios (\eg{} one scenario for if Alice says yes, and a different scenario for no), and 2) the temporal formulas for robot tasks necessarily depend on the robot skills.
We thus introduce the \emph{\roboeval{} Temporal Language} (RTL), a language derived from LTL that is particularly well-suited to specifying temporal logic formulas for robot tasks. \subfigref{codebotler-roboeval}{b} shows the grammar of RTL, and \subfigref{codebotler-roboeval}{c} shows the corresponding RTL formula for task $\mathcal{T}_1$. An additional advantage of the condition expressed in RTL \vs{} LTL is improved readability.

\subsection{The \roboeval{} Benchmark Tasks}
The \roboeval{} benchmark contains a suite of 16 tasks. \figref{benchmark-tasks} shows the names of the these tasks, along with a detailed example of the task \texttt{CountSavory} \footnote{A comprehensive list of the task descriptions can be found at \href{https://amrl.cs.utexas.edu/codebotler/}{https://amrl.cs.utexas.edu/codebotler/}
}. These tasks are designed to check whether an LMP can  
\begin{enumerate*}
    \item ground language instructions to correct function calls to robot primitives;
    \item perform accurate sequencing of robot actions;
    \item handle complex control flows based on different world configurations;
    \item solve arithmetic problems;
    \item comprehend open-world knowledge.
\end{enumerate*}
In addition, research has shown~\cite{micelibarone2023larger} that LLMs may not be as robust as previously thought, and trivial prompt variations could cause significant performance variations for LLMs~\cite{dhole2022nlaugmenter, studenteval}. For this reason, we provide 5 different paraphrases of the task prompt to evaluate the robustness of an LLM in dealing with slight prompt variations. 

\section{Benchmark Results And Analysis}

\begin{figure*}[h]
    \centering
    \resizebox{\textwidth}{!}{
        \input{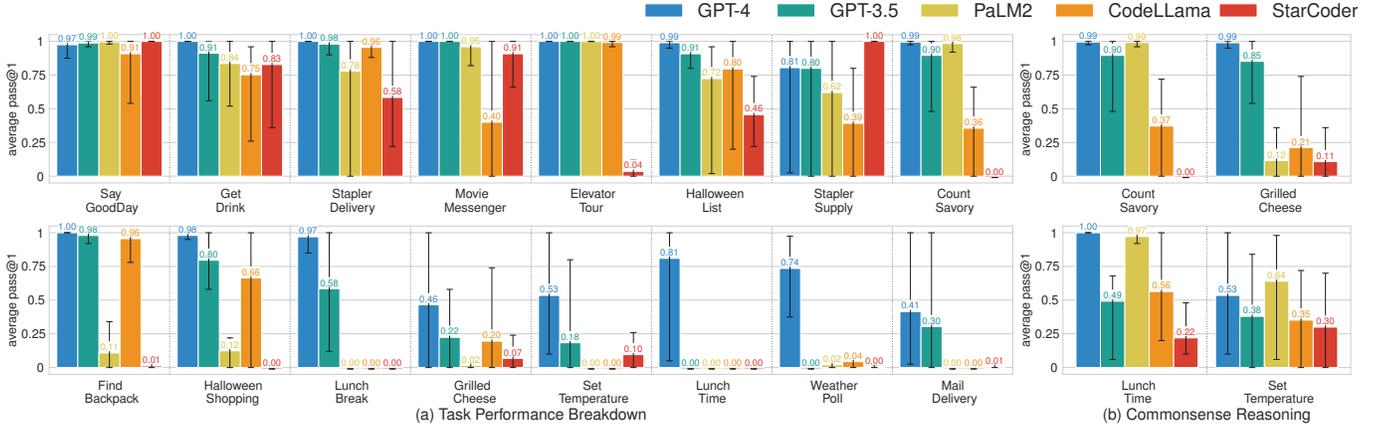}
    }
    \caption{5 LLMs are evaluated on the \roboeval{} benchmark. Each benchmark task contains 5 different prompt paraphrases and     
    each bar represents the average pass@1 score of an LLM for generating responses across all 5 prompts within a given \roboeval{} benchmark task. Each error bar indicates the range from the highest to the lowest pass@1 score across all prompts. On the left side (a), the performance of LLMs on each task of the \roboeval{} benchmark is displayed. On the right side (b), the chart shows the performance of LLMs on tasks that have been adapted to exclusively evaluate the models' proficiency in performing commonsense reasoning.}
    \label{fig:performance-bar}
    \figspace
\end{figure*}
\begin{figure}[h]
    \centering
    \resizebox{\columnwidth}{!}{
        \input{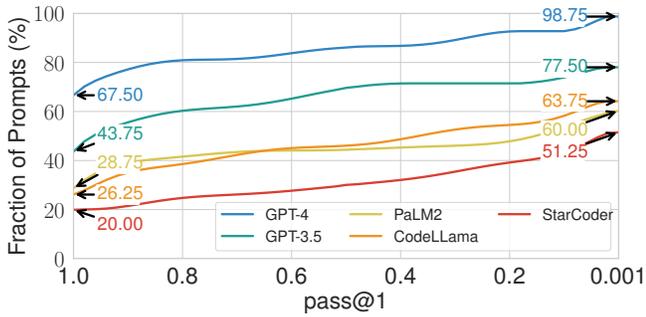}
    }
    \caption{Cumulative Distribution Function (CDF) curves depict the percentage of prompts for which each LLM can generate correct LMPs at various pass@1 score thresholds. A perfect LLM would show a horizontal line at 100\%, indicating it can generate correct LMPs for all prompts with a pass@1 score of 1. To maintain visual clarity, we limit the x-axis to $10^{-3}$ since all CDF plots eventually reach 100\%.}
    \label{fig:performance-cdf}
    \figspace
\end{figure}

\begin{figure*}[h]
    \centering
    \resizebox{\textwidth}{!}{
        \input{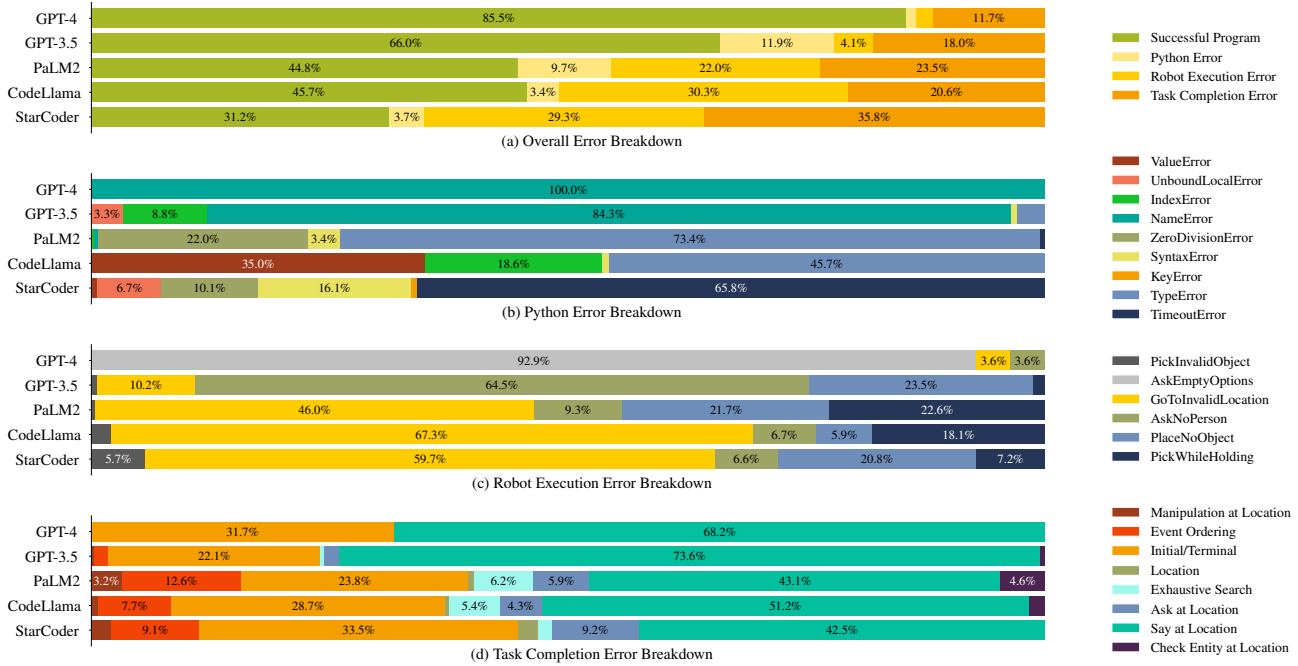}
    }
    \caption{Causes of failures for LMPs on the \roboeval{} benchmark.}
    \label{fig:cause-of-failures}
    \figspace
\end{figure*}

To gain insights into the capabilities and limitations of different state-of-the-art LLMs for generating service mobile robot LMPs, we use the \roboeval{} benchmark to empirically answer the following questions:
\begin{enumerate}
    \item First, how do different LLMs perform in generating programs for tasks in the RoboEval benchmark?
    \item Second, when a generated service robot LMP fails, what are the causes?
\end{enumerate}
To investigate these two questions, we evaluate five LLMs:
\begin{enumerate*}
    \item GPT-4~\cite{openai2023gpt4}, 
    \item GPT-3.5~\cite{brown2020language} (\texttt{text-davinci-003}), and 
    \item PaLM2~\cite{anil2023palm} (\texttt{text-bison-001}) as state-of-the-art API-only proprietary models; and 
    \item CodeLlama~\cite{roziere2023code} (\texttt{Python-34b-hf}) and 
    \item StarCoder~\cite{li2023starcoder} as open-access models.
\end{enumerate*}
When evaluating code generation models, we use standard values of $T=0.2$ for temperature and $p=0.95$ for nucleus sampling \cite{humaneval}. In the following subsections, we discuss our analysis of each question in detail.

\subsection{Performance Of LLMs On The RoboEval Benchmark}

The \roboeval{} benchmark consists of 16 tasks, each with 5 prompt paraphrases, totaling 80 different prompts. For each prompt, we generate 50 program completions and calculate the pass@1 score~\cite{humaneval}, a common metric for LMP evaluation. This score indicates the probability of an LMP being correct if an LLM generates only one LMP for a given prompt.

\emph{\textbf{Overall Performance Analysis.}}\label{overall-performance-analysis}
We first investigate the overall performance of each LLM in generating LMPs. 
Since each LLM gets a pass@1 score for every prompt, we compute the percentage of prompts that have a pass@1 score greater than or equal to a threshold value, which ranges from $1$ to $0$. We present this information in \figref{performance-cdf} as a Cumulative Distribution Function (CDF).
Although relaxing the pass@1 score threshold for each LLM increases prompt coverage, there are still certain prompts (ranging from 48.75\% for StarCoder to 1.25\% for GPT-4) where LLMs consistently fail to generate correct LMPs.


\emph{\textbf{Performance of LLMs on Individual Tasks.}}
We then look into the performance of each LLM for specific tasks. We present this information in \subfigref{performance-bar}{a}. Since each task has 5 different prompt paraphrases and a pass@1 score is calculated for each prompt, we average the pass@1 score across 5 prompts. Additionally, we plot an error bar to visualize the variation between the highest to the lowest pass@1 score over all prompts for each task. 

From the error bars, we note a considerable disparity between a model's best and worst pass@1 scores for a given task. This suggests that models are not robust to changes in the phrasing of the prompt. Single-word changes can result in substantial performance variations. A common example is changing the verb ``ask'' to ``inquire''; ``Ask him about his available ingredients'' in the \texttt{GrilledCheese} task thus becomes ``inquire about his available ingredients''.  This seems to affect some code model's ability to call the robot \texttt{ask} function.



From this chart, we further note a high variation in performance across tasks. The top row contains the high-performing tasks, where four or more models score over 0.7 on pass@1, while the bottom row contains the low-performing tasks where three or more models score below 0.2 pass@1. To identify why such a high variation exists, we run an ablation experiment. We notice that tasks involving commonsense reasoning (\texttt{CountSavory}, \texttt{GrilledCheese}, \texttt{LunchTime}, \texttt{SetTemperature}) tend to underperform. Hence, we ablate all but commonsense RTL checks on these tasks. For example, the full \texttt{CountSavory} checks require that the robot navigates to every office as well as understanding that \emph{beef jerky} and \emph{ham sandwich} are savory options; for the ablation, we remove these navigation checks. We plot the resulting average pass@1 scores of this experiment on \figref{performance-bar}{b}. If models are failing commonsense reasoning, we expect the performance to be unchanged after ablation. However, we notice that models are largely improving in performance. This suggests that the problem lies elsewhere, which we will analyze next.


\subsection{Causes of Failures of LMPs}

Given that LLMs still have room for improvement on the \roboeval{} benchmark, we want to understand the causes of failures for LLMs to generate robot programs. We classify these failures into three categories:
\begin{enumerate*}
    \item Python Errors, including syntax, runtime, and timeout errors;
    \item Robot Execution Errors, that occurs when a program attempts to execute an infeasible action, such as navigating to a non-existent (hallucinated) location; and
    \item Task Completion Errors, where the program runs correctly in the simulator but fails RTL checks for task completion.
\end{enumerate*}
We use \roboeval{}'s symbolic simulator to detect and classify Python Errors and Robot Execution Errors, and we use \roboeval{}'s evaluator to capture the Task Completion Errors. \subfigref{cause-of-failures}{a} shows the breakdown of these failure categories for each LLM. We observe that despite having fewer parameters, the CodeLLMs (CodeLlama and StarCoder) generally make fewer Python errors. This suggests that LLMs trained on a larger proportion of code may be more adept at generating successful completions in the DSL defined in the prompt. 

In the following subsections, we will analyze each error in detail.

\emph{\textbf{Python Error Analysis.}}
 \subfigref{cause-of-failures}{b} shows the specific error breakdown of the Python Error. The Python Error encompasses a multitude of errors, but their distribution exhibits a long-tailed pattern. In GPT-3.5 and GPT-4, \emph{NameError} is predominant because of undefined variables. PaLM2 and CodeLLama, on the other hand, often generate \emph{TypeError} due to the misuse of data types, while Starcoder commonly encounters \emph{TimeoutError} when the LMPs get stuck in loops.

\emph{\textbf{Robot Execution Error.}}
 \subfigref{cause-of-failures}{c} shows the error breakdown of the Robot Execution Error. There are 6 root causes  of robot execution errors: 
 \begin{enumerate}
     \item GoToInvalidLocation/PickInvalidObject: the program calls \texttt{go\_to} or \texttt{pick} with a hallucinated argument;
     \item PlaceNoObject/PickWhileHolding: the program tries to pick/place an object when it is not/already holding one;
     \item AskNoPerson: the program calls \texttt{ask} at a location with no person nearby; and 
     \item AskEmptyOptions: the program calls \texttt{ask} with an empty list of options for the person to choose from.
 \end{enumerate}   
 We first observe that hallucination plays a substantial role in causing errors, Specifically, the GoToInvalidLocation and PickInvalidObject errors contribute to 33.7\% of the total robot execution errors in GPT-3.5, 67.7\% in PaLM2, 73.2\% in CodeLlama, and 80.5\% in StarCoder. We also notice that another important source of errors arises from the PlaceNoObject and PickWhileHolding errors, as well as AskNoPerson errors for GPT-3.5. These errors require the program to be aware of the internal state of the robot or keep track of the external world state. The prevalence of these errors suggests a gap in LLM's ability to keep track of changing states by binding agents to current states. Some of these errors also result from a failure to respect the behavior of our robot functions. For example, our \texttt{pick} function only allows the robot to pick an object if it is not already holding one. However, the model may not include this information in its completion and attempt to pick up multiple objects.
 
 \begin{figure*}[t]
    \begin{minipage}{0.63\textwidth}
        \centering
        \resizebox{\linewidth}{!}{\input{figures/resample_cdf.pgf}}
        \caption{Cumulative Distribution Function (CDF) of the LLMs' \\
        performance across different max retry limits. As the max retry \\
        limit increases, all five LLMs improve in performance.}
        \label{fig:cdf-after-resampling}
    \end{minipage}%
    \hspace{-1cm}
    \begin{minipage}{0.4\textwidth}
        \centering
        \resizebox{\linewidth}{!}{\input{figures/resample_till_no_crash.pgf}}
        \caption{Cumulative Distribution Function (CDF) of the fraction of program completions that can be executed over different max retry limits.}
        \label{fig:resample-till-no-crash}
    \end{minipage}
    \figspace
\end{figure*}
 
\emph{\textbf{Task Completion Error.}} 
\subfigref{cause-of-failures}{d} shows the breakdown of the Task Completion Errors. We classify every temporal check in the \roboeval{} benchmark into one of the following categories:
 \begin{enumerate}
     \item \{Say/Ask/Manipulation/CheckEntity\} AtLocation: The task requires executing a specific action (\texttt{say}, \texttt{ask}, \texttt{pick}, \texttt{place} or \texttt{check}) at a specific location, but the program fails to do so;
     \item Initial/Terminal: The program does not accurately perform an initial or final action;
     \item EventOrdering: The program does not carry out actions in the prescribed sequence or has redundant navigation;
     \item Location: The program commands the robot to visit a location irrelevant to the task; and, 
     \item ExhaustiveSearch: The program does not visit all locations required to complete the task.
 \end{enumerate}
 We notice that the most common task completion errors in each LLM are the SayAtLocation error and the Initial/Terminal error. These two types of errors constitute the majority of task completion errors, making up $100\%$ of the errors in GPT-4, $95.7\%$ in GPT-3.5, $66.9\%$ in PaLM2, $79.9\%$ in CodeLlama, and $76.0\%$ in StarCoder.

\begin{figure}[htp]
\includegraphics[clip,width=\columnwidth]{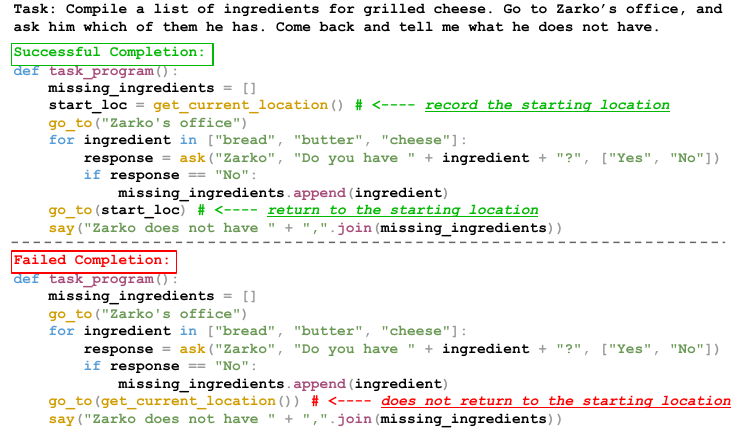}%
\caption{Examples of successful (top) and failed (bottom) program completions related to the \emph{``come back''} instruction. }
\label{fig:comeback}
\figspace
\end{figure}
After analyzing the programs that produce these errors, we find that most are related to a recurring statement in many of our tasks: \emph{``Come back and tell me..."} (\figref{comeback} shows an example). This statement is often used in tasks in which we want a service robot to go to a different location, complete a task, and then return to us with an update on its progress. This is the most common instruction that fails across models, and also the error that is directly responsible for many of our low-performing tasks. Programs that fail the \emph{``come back''} instruction often do not use the \texttt{get\_current\_location} primitive to record the robot's starting location, and as a result, cannot refer to the starting location at the end of the program. This also causes LLMs to hallucinate locations and produce the GoToInvalidLocation errors --- rather than referencing a non-existent \texttt{start\_loc} variable, some LLMs generate a location based on context clues and send the robot to that location (for example, returning to the kitchen in \texttt{GrilledCheese}, although no kitchen is mentioned in the prompt). We observe that in tasks without a \emph{``come back''} instruction, but an explicit return statement like \emph{``return to the mail room''} do not suffer from the same error.

\label{sec:results}

\section{Improving  Robot Program Generations}
 Based on the analysis of the failures of LMPs in \roboeval{}, we are interested in understanding how to improve service robot program generation using LLMs. Recognizing the breadth of potential improvements, this study focuses on an initial step: we propose a \emph{rejection sampling strategy} to identify and reduce LMP failures (Python Errors and Robot Execution Errors) before deploying the LMP on the robot.

 To detect errors in an LMP, the \roboeval{} symbolic simulator takes in the current world state and executes the LMP. If an error is identified (Python Errors and Robot Execution Errors), \codebotler{} will prompt an LLM for a new program and submit the program to the symbolic simulator to execute again. This cycle repeats until an LMP successfully passes in the symbolic simulator for deployment on the robot or until a maximum retry limit is reached.

 
 This proposed strategy has one limitation: the symbolic simulator may not know a-priori the true state of the world, including the current locations of humans and movable objects, or how humans might respond to the robot's questions. We address this limitation by proposing a task-agnostic world state. This world state contains the permanent entities (\eg{} known rooms) and employs \emph{state sampling} to simulate random potential world states for non-static entities, such as possible human locations, movable objects, and human responses. Subsequently, each LMP undergoes multiple simulation runs (we chose 5 in our experiments) in the symbolic simulator to ensure statistical reliability when identifying LMP failures.

 We evaluate this strategy on all five LLMs with four different maximum retry limits ($2,4,8,100$) and compare them with the baseline (without rejection sampling). \figref{cdf-after-resampling} shows the CDF curves of the percentage of prompts that can be successfully generated given a threshold of the pass@1 score for each LLM with respect to different maximum retry limits. We observe an improvement in performance across all LLMs as the maximum retry limit is increased.

  We then investigate how effective the rejection sampling strategy is in eliminating the program execution errors. We compute the percentage of total program completions that can be eventually executed over different maximum retry limits and plot it as a CDF in \figref{resample-till-no-crash}. Interestingly, we observe that it is possible for some LLMs (PaLM2, CodeLlama, and Starcoder) to never generate successful completions for certain tasks. As a result, while the rejection sampling strategy can improve the performances of LLMs, it is not enough to resolve all program execution errors. 
 
This observation, coupled with prior findings of LLMs' consistent failures in generating correct LMPs as detailed in Section~\ref{overall-performance-analysis}, points to a systemic challenge in LMP generation. It highlights the issue of abstraction matching~\cite{Liu_2023}, which entails aligning ambiguous natural language expressions of user intent with their precise, unambiguous code representations. In future work, we would like to explore more sophisticated strategies, such as utilizing grounded abstraction matching, to solve this problem.


\section{Conclusion, Limitations, \& Future Works}
In this work, we present \codebotler{} and \roboeval{}. \codebotler{} is an open-source robot-agnostic tool to generate general-purpose service robot programs from natural language, and \roboeval{} is a benchmark for evaluating LLMs’ capabilities of generating programs to complete service robot tasks. We evaluate the performance of five LLMs in generating robot programs and perform an analysis of the causes of failures. Our analysis reveals that the errors exhibit a long-tail distribution, with LLMs predominantly struggling with hallucination issues and grounding the phrase \emph{``come back"}. Finally, we propose a rejection sampling strategy to handle program failures. This method has led to improved performance for all five LLMs.

This work has several limitations that could be addressed in future research. Firstly, \codebotler{} currently does not support low-level behaviors, such as \emph{``follow Alice to her office''}. Secondly, \codebotler{} generates LMPs in an open-loop fashion, rendering it incapable of reacting to unexpected changes in the environment. Thirdly, this study does not consider the strategies for crafting prompts that could improve the performance of LLMs in generating service robot programs. Finally, although the RTL constraints are designed to reduce the workloads of writing specifications compared to LTL constraints, the users still need to manually specify constraints for each task. Therefore, investigating the Tree-of-Thoughts concept~\cite{yao2023tree} to dynamically generate RTL checks with LLMs might be valuable.


\bibliographystyle{IEEEtran}
\bibliography{references}

\begin{thebibliography}{10}
\providecommand{\url}[1]{#1}
\csname url@rmstyle\endcsname
\providecommand{\newblock}{\relax}
\providecommand{\bibinfo}[2]{#2}
\providecommand\BIBentrySTDinterwordspacing{\spaceskip=0pt\relax}
\providecommand\BIBentryALTinterwordstretchfactor{4}
\providecommand\BIBentryALTinterwordspacing{\spaceskip=\fontdimen2\font plus
\BIBentryALTinterwordstretchfactor\fontdimen3\font minus \fontdimen4\font\relax}
\providecommand\BIBforeignlanguage[2]{{%
\expandafter\ifx\csname l@#1\endcsname\relax
\typeout{** WARNING: IEEEtran.bst: No hyphenation pattern has been}%
\typeout{** loaded for the language `#1'. Using the pattern for}%
\typeout{** the default language instead.}%
\else
\language=\csname l@#1\endcsname
\fi
#2}}

\bibitem{codeaspolicies2022}
J.~Liang, W.~Huang, F.~Xia, P.~Xu, K.~Hausman, B.~Ichter, P.~Florence, and A.~Zeng, ``{Code as Policies: Language Model Programs for Embodied Control},'' in \emph{arXiv:2209.07753}, 2022.

\bibitem{singh2023progprompt}
I.~Singh, V.~Blukis, \emph{et~al.}, ``{ProgPrompt: Generating Situated Robot Task Plans using Large Language Models},'' in \emph{ICRA 2023}, 2023.

\bibitem{ahn2022isaycan}
M.~Ahn, A.~Brohan, \emph{et~al.}, ``{Do As I Can, Not As I Say: Grounding Language in Robotic Affordances},'' 2022.

\bibitem{huang2022language}
W.~Huang, P.~Abbeel, D.~Pathak, and I.~Mordatch, ``{Language Models as Zero-Shot Planners: Extracting Actionable Knowledge for Embodied Agents},'' \emph{International Conference on Learning Representations}, 2022.

\bibitem{driess2023palme}
D.~Driess, F.~Xia, \emph{et~al.}, ``{PaLM-E: An Embodied Multimodal Language Model},'' \emph{Proceedings of the 40th International Conference on Machine Learning (ICML), no. 340, pp. 8469–8488}, Jul. 2023.

\bibitem{liu2023llmp}
B.~Liu, Y.~Jiang, \emph{et~al.}, ``{LLM+P: Empowering Large Language Models with Optimal Planning Proficiency},'' \emph{arXiv:2304.11477}, 2023.

\bibitem{rosactionlib}
\BIBentryALTinterwordspacing
``{ROS Actionlib},'' accessed 2023-10-19. [Online]. Available: \url{http://wiki.ros.org/actionlib}
\BIBentrySTDinterwordspacing

\bibitem{humaneval}
M.~Chen, J.~Tworek, H.~Jun, Q.~Yuan, \emph{et~al.}, ``{Evaluating large language models trained on code},'' \emph{arXiv:2107.03374}, 2021.

\bibitem{studenteval}
H.~M. Babe, S.~Nguyen, Y.~Zi, A.~Guha, M.~Q. Feldman, and C.~J. Anderson, ``{StudentEval}: A benchmark of student-written prompts for large language models of code,'' \emph{arXiv:2306.04556}, 2023.

\bibitem{huang2023vlmap}
C.~Huang, O.~Mees, A.~Zeng, and W.~Burgard, ``{Visual Language Maps for Robot Navigation},'' \emph{IEEE International Conference on Robotics and Automation (ICRA), pp. 10608-10615}, 2022.

\bibitem{ding2023task}
Y.~Ding, X.~Zhang, C.~Paxton, and S.~Zhang, ``{Task and Motion Planning with Large Language Models for Object Rearrangement},'' \emph{IEEE/RSJ International Conference on Intelligent Robots and Systems (IROS), pp. 2086-2092}, 2023.

\bibitem{wang2023demo2code}
H.~Wang, G.~Gonzalez-Pumariega, Y.~Sharma, and S.~Choudhury, ``{Demo2Code: From Summarizing Demonstrations to Synthesizing Code via Extended Chain-of-Thought},'' \emph{37th Conference on Neural Information Processing Systems}, 2023.

\bibitem{huang2023voxposer}
W.~Huang, C.~Wang, \emph{et~al.}, ``{VoxPoser: Composable 3D Value Maps for Robotic Manipulation with Language Models},'' \emph{Proceedings of The 7th Conference on Robot Learning, PMLR vol. 229, pp. 540-562}, 2023.

\bibitem{brohan2023rt2}
A.~Brohan, N.~Brown, \emph{et~al.}, ``{RT-2: Vision-Language-Action Models Transfer Web Knowledge to Robotic Control},'' \emph{Proceedings of The 7th Conference on Robot Learning, PMLR vol. 229, pp. 2165--2183}, 2023.

\bibitem{tang2023saytap}
Y.~Tang, W.~Yu, \emph{et~al.}, ``{SayTap: Language to Quadrupedal Locomotion},'' \emph{Proceedings of The 7th Conference on Robot Learning, PMLR vol. 229, pp. 3556-3570}, 2023.

\bibitem{yu2023language}
W.~Yu, N.~Gileadi, C.~Fu, \emph{et~al.}, ``{Language to Rewards for Robotic Skill Synthesis},'' \emph{Proceedings of The 7th Conference on Robot Learning, PMLR vol. 229, pp. 374-404}, 2023.

\bibitem{shah2022lmnav}
D.~Shah \emph{et~al.}, ``{LM-Nav: Robotic Navigation with Large Pre-Trained Models of Language, Vision, and Action},'' \emph{Proceedings of The 6th Conference on Robot Learning, PMLR vol. 205, pp. 492-504}, 2022.

\bibitem{huang2022inner}
W.~Huang, F.~Xia, \emph{et~al.}, ``{Inner Monologue: Embodied Reasoning through Planning with Language Models},'' \emph{Proceedings of The 6th Conference on Robot Learning, PMLR vol. 205, pp. 1769-1782}, 2022.

\bibitem{wu2023tidybot}
J.~Wu, R.~Antonova, \emph{et~al.}, ``{TidyBot: Personalized Robot Assistance with Large Language Models},'' \emph{Autonomous Robots, vol. 47, no. 8, pp. 1087–1102}, 2023.

\bibitem{chen2022openvocabulary}
B.~Chen, F.~Xia, \emph{et~al.}, ``{Open-vocabulary Queryable Scene Representations for Real World Planning},'' \emph{IEEE International Conference on Robotics and Automation (ICRA), pp. 11509-11522}, 2022.

\bibitem{song2023llmplanner}
C.~H. Song, J.~Wu, \emph{et~al.}, ``{LLM-Planner: Few-Shot Grounded Planning for Embodied Agents with Large Language Models},'' \emph{Proceedings of IEEE/CVF International Conference on Computer Vision (ICCV)}, 2023.

\bibitem{brown2020language}
T.~Brown, B.~Mann, N.~Ryder, \emph{et~al.}, ``{Language Models are Few-Shot Learners},'' in \emph{Advances in Neural Information Processing Systems}, vol.~33, 2020, pp. 1877--1901.

\bibitem{chang2023survey}
Y.~Chang \emph{et~al.}, ``A survey on evaluation of large language models,'' \emph{ACM Transactions on Intelligent Systems and Technology}, 2024.

\bibitem{liang2023holistic}
P.~Liang, R.~Bommasani, T.~Lee, \emph{et~al.}, ``{Holistic Evaluation of Language Models},'' \emph{Transactions on Machine Learning Research}, 2023.

\bibitem{liu2023agentbench}
X.~Liu, H.~Yu, \emph{et~al.}, ``{AgentBench: Evaluating LLMs as Agents},'' \emph{The 12th International Conference on Learning Representations}, 2023.

\bibitem{lai2022ds1000}
Y.~Lai \emph{et~al.}, ``{DS-1000: A Natural and Reliable Benchmark for Data Science Code Generation},'' \emph{Proceedings of the 40th International Conference on Machine Learning, PMLR vol. 202, pp.18319-18345}, 2023.

\bibitem{austin2021program}
J.~Austin, A.~Odena, \emph{et~al.}, ``{Program Synthesis with Large Language Models},'' \emph{in arXiv:2108.07732}, 2021.

\bibitem{OpenAIPlatform}
\BIBentryALTinterwordspacing
``{OpenAI Platform},'' accessed: 2023-9-10. [Online]. Available: \url{https://platform.openai.com/docs/guides/gpt}
\BIBentrySTDinterwordspacing

\bibitem{PaLMAPI}
\BIBentryALTinterwordspacing
``{Generative AI for Developers},'' accessed: 2023-9-10. [Online]. Available: \url{https://developers.generativeai.google/}
\BIBentrySTDinterwordspacing

\bibitem{micelibarone2023larger}
A.~V. Miceli-Barone, F.~Barez, I.~Konstas, and S.~B. Cohen, ``{The Larger They Are, the Harder They Fail: Language Models do not Recognize Identifier Swaps in Python},'' \emph{61st Annual Meeting of the Association for Computational Linguistics}, 2023.

\bibitem{dhole2022nlaugmenter}
K.~D. Dhole, V.~Gangal, S.~Gehrmann, \emph{et~al.}, ``{NL-Augmenter: A Framework for Task-Sensitive Natural Language Augmentation},'' \emph{Northern European Journal of Language Technology}, 2021.

\bibitem{openai2023gpt4}
OpenAI, ``{GPT-4 Technical Report},'' \emph{arXiv:2303.08774}, 2023.

\bibitem{anil2023palm}
R.~Anil, A.~M. Dai, O.~Firat, \emph{et~al.}, ``{PaLM 2 Technical Report},'' \emph{arXiv:2305.10403}, 2023.

\bibitem{roziere2023code}
B.~Rozière, J.~Gehring, F.~Gloeckle, \emph{et~al.}, ``{Code Llama: Open Foundation Models for Code},'' \emph{arXiv:2308.12950}, 2023.

\bibitem{li2023starcoder}
R.~Li, L.~B. Allal, Y.~Zi, \emph{et~al.}, ``{StarCoder: may the source be with you!}'' \emph{Transactions on Machine Learning Research}, 2023.

\bibitem{Liu_2023}
M.~X. Liu, A.~Sarkar, \emph{et~al.}, ``{\textquotedblleft}what it wants me to say{\textquotedblright}: Bridging the abstraction gap between end-user programmers and code-generating large language models,'' in \emph{Proceedings of the 2023 {CHI} Conference on Human Factors in Computing Systems}, 2023.

\bibitem{yao2023tree}
S.~Yao, D.~Yu, \emph{et~al.}, ``{Tree of Thoughts: Deliberate Problem Solving with Large Language Models},'' \emph{Proceedings of the 37th International Conference on Neural Information Processing Systems}, 2023.

\end{thebibliography}

\end{document}